\documentclass{article}

\usepackage[preprint]{neurips_2025}

\usepackage[utf8]{inputenc} 
\usepackage[T1]{fontenc}    
\usepackage{hyperref}       
\usepackage{url}            
\usepackage{booktabs}       
\usepackage{amsfonts}       
\usepackage{nicefrac}       
\usepackage{microtype}      
\usepackage{xcolor, epsfig}         
\usepackage{amsmath, amsthm, amssymb, lipsum, multicol}
\usepackage{algorithm}

\title{Joint Tensor-Train Parameterization for Efficient and Expressive Low-Rank Adaptation}

\author{ 
Jun Qi \\
Georgia Institute of Technology\\
jqi41@gatech.edu
\And
Chen-Yu Liu \\
National Taiwan University \\
d10245003@g.ntu.edu.tw \\
\And
Sabato Marco Siniscalchi  \\
University of Palermo  \\
sabatomarco.siniscalchi@unipa.it
\And
Chao-Han Huck Yang\\
NVIDIA Research\\
hucky@nvidia.com 
\And
Min-Hsiu Hsieh\\
Hon-Hai Quantum Computing Research Center \\
min-hsiu.hsieh@Foxconn.com
}

\begin{document}

\maketitle

\begin{abstract}
Low-Rank Adaptation (LoRA) is widely recognized for its parameter-efficient fine-tuning of large-scale neural models. However, standard LoRA independently optimizes low-rank matrices, which inherently limits its expressivity and generalization capabilities. While classical tensor-train (TT) decomposition can be separately employed on individual LoRA matrices, this work demonstrates that the classical TT-based approach neither significantly improves parameter efficiency nor achieves substantial performance gains. This paper proposes TensorGuide, a novel tensor-train-guided adaptation framework to overcome these limitations. TensorGuide generates two correlated low-rank LoRA matrices through a unified TT structure driven by controlled Gaussian noise. The resulting joint TT representation inherently provides structured, low-rank adaptations, significantly enhancing expressivity, generalization, and parameter efficiency without increasing the number of trainable parameters. Theoretically, we justify these improvements through neural tangent kernel analyses, demonstrating superior optimization dynamics and enhanced generalization. Extensive experiments on quantum dot classification and GPT-2 fine-tuning benchmarks demonstrate that TensorGuide-based LoRA consistently outperforms standard LoRA and TT-LoRA, achieving improved accuracy and scalability with fewer parameters.
\end{abstract}

\section{Introduction}

Fine-tuning large-scale pre-trained models has become standard practice across diverse machine learning tasks, from natural language processing~\cite{deng2018deep} to quantum computing~\cite{caro2022generalization}. Despite their widespread success, these models present significant computational and memory overhead, motivating the development of parameter-efficient fine-tuning methods, such as Low-Rank Adaptation (LoRA)~\cite{hu2022lora}. LoRA introduces lightweight, trainable low-rank matrices that preserve most of the original pre-trained weights, enabling efficient task-specific adaptation. However, the traditional LoRA framework has two critical shortcomings: first, the separately parameterized low-rank matrices inherently limit expressive capability and induce poor conditioning~\cite{zeng2023expressive}; second, the chosen low-rank dimensionality severely constrains the balance between efficiency and generalization, restricting model performance~\cite{liu2024dora}.

\begin{figure}
\centerline{\includegraphics[width=4.5in]{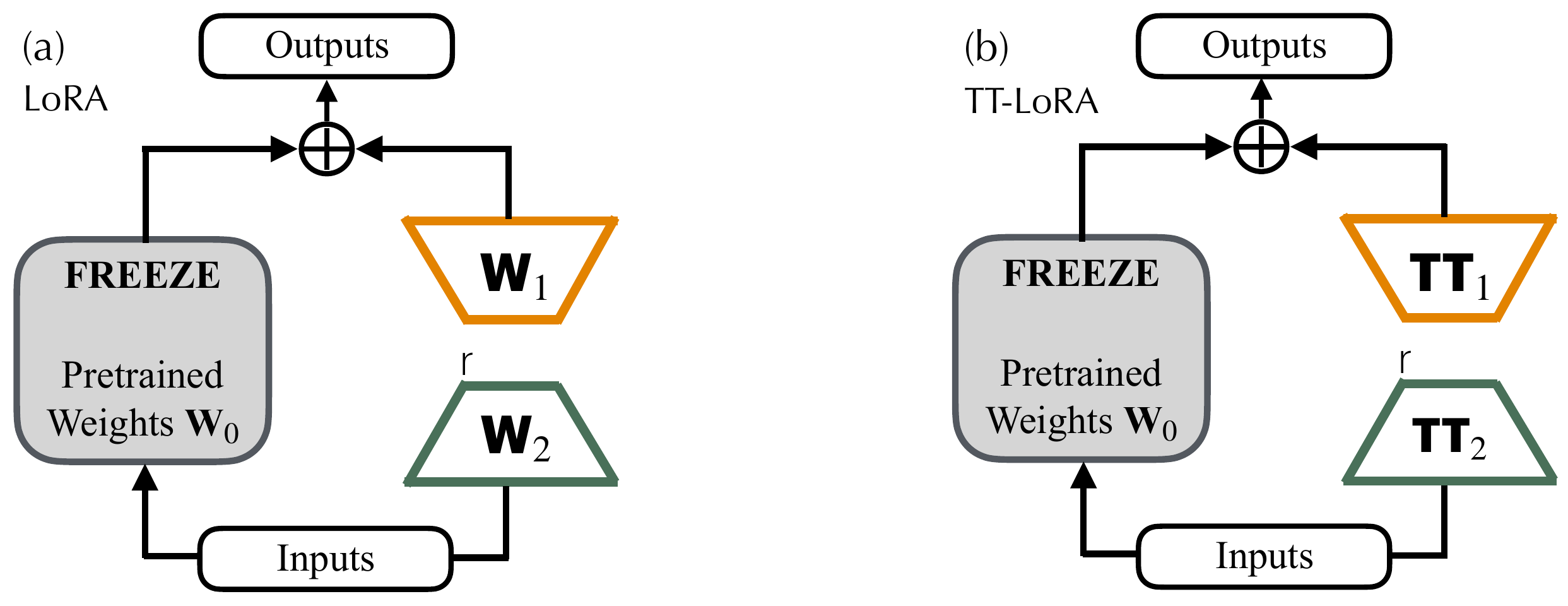}} 
\caption{{\it Comparison between (a) standard LoRA and (b) TT-LoRA}. Both methods freeze the pretrained weight matrix $\textbf{W}_{0}$ and inject trainable low-rank adaptation. In (a), LoRA directly optimizes two low-rank matrices $\textbf{W}_{1} \in \mathbb{R}^{D\times r}$ and $\textbf{W}_{2} \in \mathbb{R}^{r\times Q}$, where $r$ is the intrinsic rank. In (b), TT-LoRA replaces each low-rank matrix with a TT decomposition, namely $\rm TT_{1}$ and $\rm TT_{2}$. While TT-LoRA introduces structured factorization, it does not increase expressivity beyond standard LoRA for the same rank $r$, nor significantly reduce parameter count due to separate decomposition $\textbf{W}_{1}$ and $\textbf{W}_{2}$.}
\label{fig:lora}
\end{figure}

As shown in Figure~\ref{fig:lora}, Tensor-Train LoRA (TT-LoRA) could address these limitations, decomposing each LoRA adaptation matrix independently into structured tensor-train formats~\cite{oseledets2011tensor}. However, TT-LoRA does not significantly enhance the expressivity or parameter efficiency since decomposing each low-rank matrix separately yields negligible reductions in trainable parameters and does not introduce beneficial inter-matrix correlations~\cite{zaken2021bitfit}.

In contrast, our proposed framework, TensorGuide, utilizes a single tensor-train (TT) network to jointly generate both low-rank matrices. As shown in Figure~\ref{fig:tt2lora}, by feeding the TT network structured Gaussian noise, TensorGuide creates correlated parameter adaptations, which dramatically enhance expressivity and conditioning without increasing the parameter count. Notably, the structured low-rank TT representation enables scaling the hidden layer width of the adaptation network arbitrarily, thereby decoupling model expressivity from parameter cost.

We theoretically support our claims using neural tangent kernel (NTK)~\cite{jacot2018neural, bietti2019inductive} analysis and eigenvalue bounds, demonstrating that TensorGuide achieves superior generalization and convergence rates compared to standard LoRA and TT-LoRA. Intuitively, NTK captures how small changes in a model's parameters affect its output during gradient descent, effectively characterizing the network's function space as a kernel method. A larger minimum eigenvalue of the NTK implies a smoother loss landscape, improved trainability, and stronger generalization. In the context of TensorGuide, we demonstrate that the structured, low-rank nature of TensorGuide yields an NTK with superior spectral properties compared to traditional LoRA baselines. This connection helps justify our observed empirical improvements in both optimization and generalization. 

Moreover, extensive experimental evaluations in quantum computing and natural language processing tasks confirm the robustness and scalability of TensorGuide. Our method consistently outperforms established parameter-efficient fine-tuning baselines across classical and quantum computing domains. Our theoretical analysis and empirical results collectively illustrate that TensorGuide substantially advances parameter-efficient fine-tuning methodologies~\cite{ding2023parameter, liu2022few, yang2021voice2series}, effectively combining theoretical rigor with practical efficacy.



\section{Background and Related Work}

\subsection{Low-Rank Adaptation} 

In recent years, we have witnessed the rapid proliferation of large-scale pre-trained models, including transformer-based architectures such as GPT~\cite{radford2019language, brown2020language} and Vision Transformers~\cite{dosovitskiy2020image}. Fine-tuning these models on downstream tasks is computationally expensive due to the extensive number of parameters. LoRA has been proposed as an efficient approach for fine-tuning to alleviate this challenge. LoRA introduces task-specific low-rank matrices to modulate the frozen pre-trained weights, thereby dramatically reducing the number of trainable parameters. Specifically, given a pre-trained weight matrix $\textbf{W}_0 \in \mathbb{R}^{D \times Q}$, LoRA parameterizes the adapted weights as: 
\begin{equation}
\textbf{W} = \textbf{W}_0 + \Delta \textbf{W}, \quad{\rm where} \hspace{1mm} \Delta \textbf{W} = \textbf{W}_2 \textbf{W}_1, \hspace{1mm} \textbf{W}_1 \in \mathbb{R}^{D \times r}, \hspace{1mm}  \textbf{W}_2 \in \mathbb{R}^{r \times Q},
\end{equation}
with $r \ll \min(D, Q)$ typically representing the rank. Despite its popularity and effectiveness, standard LoRA faces inherent limitations in expressivity and generalization, particularly in high-dimensional adaptation scenarios. Moreover, its parameter efficiency is strictly tied to the choice of the adaptation rank, limiting potential scalability and adaptability to larger models. 

\subsection{Tensor-Train Decomposition}

The tensor-train (TT) decomposition~\cite{oseledets2011tensor} is a powerful tensor factorization method designed to compress high-dimensional tensors efficiently~\cite{orus2014practical}. Given an order-$K$ tensor $\mathcal{W} \in \mathbb{R}^{d_1 \times d_2 \times \dots \times d_K}$, the TT decomposition represents it as a chain of low-dimensional cores: 
\begin{equation}
\mathcal{W}(i_1, i_2, \dots, i_K) = \mathcal{G}_1(i_1) \mathcal{G}_2(i_2) \dots \mathcal{G}_K(i_K),
\end{equation}
where each core $\mathcal{G}_{k}(i_{k})$ is a matrix whose dimensions depend on predetermined TT-ranks $(r_0, r_1, \dots, r_K)$, with $r_0=r_K=1$. TT decomposition reduces parameter complexity from exponential to linear in dimension, making it highly efficient for representing structured, low-rank tensors. TT decomposition has successfully been employed in neural network compression~\cite{schuch2008simulation, novikov2015tensorizing} and efficient parameterization for diverse machine learning tasks~\cite{yang2017tensor, qi2023exploiting, su2020convolutional}.

 \subsection{Classical TT-based Approach for Low-Rank Adaptation (TT-LoRA)}
 
Motivated by these efficiency gains, a natural extension is to separately apply classical TT decomposition to each low-rank matrix within LoRA, referred to as TT-LoRA in this work. Given $\textbf{W}_1$ and $\textbf{W}_2$ from LoRA, TT-LoRA independently decomposes each into TT format as $\rm TT_1$ and $\rm TT_2$. While such an approach theoretically reduces storage complexity and may introduce structured regularization, it does not yield a substantial reduction in parameters. Furthermore, by decomposing each matrix independently, TT-LoRA does not significantly improve the representation capability compared to the original LoRA. Its separate decomposition approach fails to exploit beneficial correlations between the adaptation matrices, ultimately restricting potential gains in generalization performance. 
 
 \subsection{Neural Tangent Kernel (NTK) Theory}
 
The NTK theory, first introduced in~\cite{jacot2018neural}, provides profound insights into the training dynamics of wide neural networks and their generalization properties. NTK theory characterizes the evolution of neural network outputs under gradient descent in the infinite-width limit, demonstrating that a linearized model around the initialization point can accurately describe model predictions during training. 
 
The NTK framework has shed light on several critical aspects of deep learning theory, including convergence guarantees, optimization landscapes, and generalization performance. For instance, NTK eigenvalue analysis can reveal conditions under which optimization becomes easier or more stable~\cite{du2019gradient, arora2019exact}. Specifically, a well-conditioned NTK, characterized by relatively large minimum eigenvalues, corresponds to stable and fast optimization, as well as improved generalization capability. 

In this work, NTK theory provides essential theoretical support for understanding and analyzing our proposed TensorGuide architecture. By leveraging structured TT parameterizations, we aim to positively influence the NTK's spectral properties, enhancing network trainability and generalization performance.

\begin{figure}
\centerline{\includegraphics[width=4.5in]{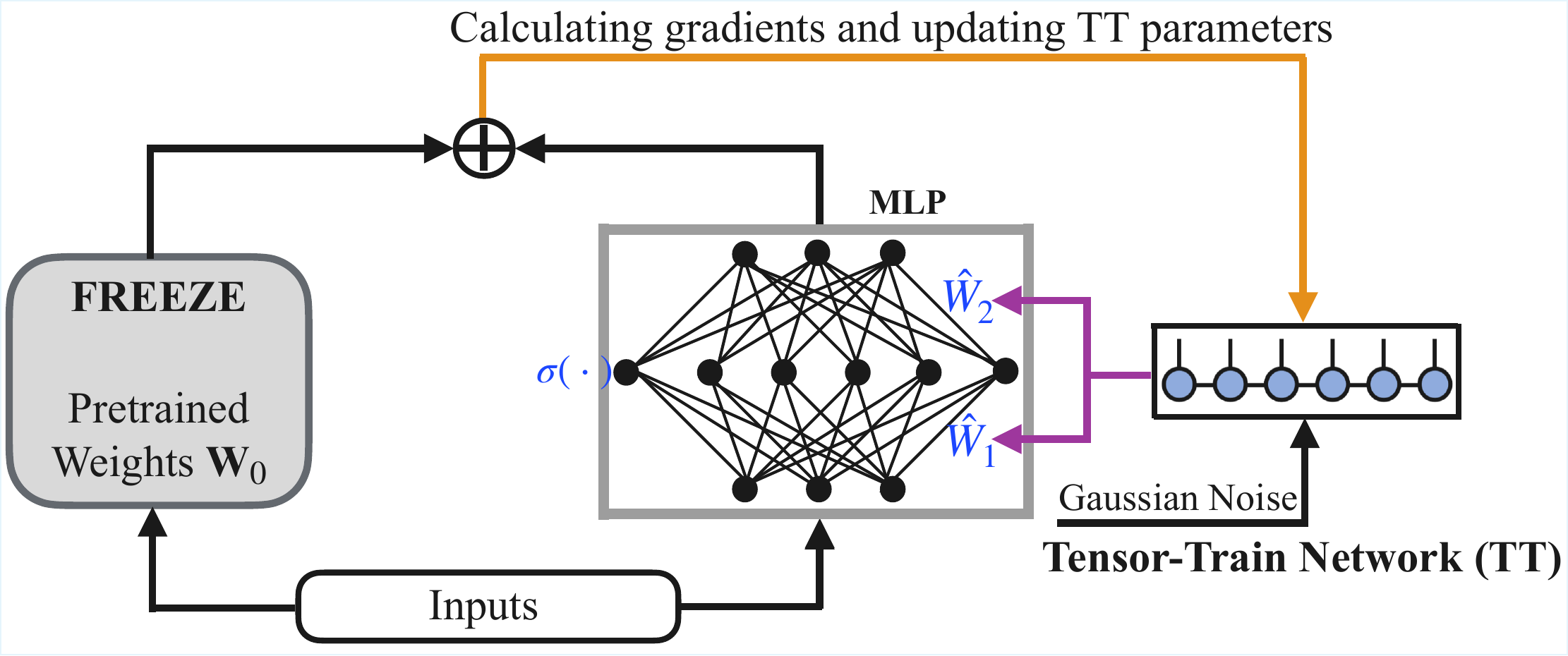}} 
\caption{{\it Overview of TensorGuide Framework}. Given frozen pre-trained weights $\textbf{W}_{0}$, the TensorGuide framework injects adaptation by generating two low-rank matrices $\hat{\textbf{W}}_{1}$ and $\hat{\textbf{W}}_{2}$ via a shared TT network, which is driven by Gaussian noise. The generated matrices modulate the output of an intermediate MLP layer $\sigma(\cdot)$. Importantly, TensorGuide enables expressive adaptation with significantly fewer trainable parameters, as the TT network inherently models structured low-rank transformations that scale efficiently with width.}
\label{fig:tt2lora}
\end{figure}

\section{TensorGuide: Architecture and Methodology}
In this section, we introduce the architecture and methodology of the proposed TensorGuide framework, as illustrated in Figure~\ref{fig:tt2lora}. TensorGuide leverages a unified TT network to generate two low-rank matrices simultaneously, denoted as $\mathbf{\hat{W}}_1$ and $\mathbf{\hat{W}}_2$, within a LoRA paradigm. Unlike standard LoRA, where low-rank matrices are optimized independently, and TT-LoRA, where each low-rank matrix is decomposed separately using TT decomposition, our approach employs a single TT network that jointly generates these matrices, significantly enhancing parameter efficiency and expressivity. 

\subsection{Unified TT Generation of LoRA Matrices}
In TensorGuide, we propose a structured approach that encodes both adaptation matrices ($\mathbf{\hat{W}}_1 \in \mathbb{R}^{D \times r}$ and $\mathbf{\hat{W}}_2 \in \mathbb{R}^{r \times Q}$) using a single TT network fed with controlled Gaussian noise. Specifically, we sample a Gaussian latent vector $\mathbf{z} \sim \mathcal{N}(0, I)$, structured in a tensorized format, and use the TT network to generate a vectorized representation of the two matrices. This single-step generation naturally couples $\mathbf{\hat{W}}_1$ and $\mathbf{\hat{W}}_2$, promoting beneficial correlations that enhance representation and generalization performance.

Formally, given TT cores $\{{\mathcal{G}_k}\}_{k=1}^{K}$, the TT network transforms a structured Gaussian latent input $\mathbf{z}$ into a concatenated vector representation of both $\mathbf{\hat{W}}_1$ and $\mathbf{\hat{W}}_2$: 
\begin{equation}
[\hat{\textbf{W}}_{1}, \hat{\textbf{W}}_{2}] = \rm TT(\textbf{z}; \{\mathcal{G}_{k}\}_{k=1}^{K}).  
\end{equation}

This approach efficiently captures the joint structure of adaptation matrices, leading to significant parameter reductions compared to standard independent TT decompositions.

\subsection{Efficient Scaling with Structured Low-Rank TT}
A key advantage of TensorGuide stems from its inherent structured low-rank tensor representation. TT decomposition enables the compact modeling of large parameter spaces, where the number of trainable parameters scales linearly with the tensor ranks and dimensions. By carefully selecting the TT-rank and the dimensions of TT cores, our method efficiently scales the practical hidden dimension of the generated MLP weights without increasing the number of trainable parameters.

To illustrate, assume that standard LoRA has intrinsic rank $r$, resulting in two low-rank matrices of total size $\mathcal{O}((D + Q) \times r)$. TT-LoRA individually decomposes each matrix, but the resulting parameter savings become negligible. Conversely, TensorGuide employs a single TT representation that is significantly smaller than $r$, thereby compressing the parameter space. This approach enables the arbitrary scaling of the hidden dimensions of the MLP, ensuring both representation capability and optimization performance, as guaranteed by the NTK theory, particularly when the MLP's hidden layer has a large width. 

\subsection{Algorithmic Implementation}

The training of TensorGuide involves optimizing the TT cores parameters $\boldsymbol{\theta} = \{\mathcal{G}_{k}\}_{k=1}^{K}$, which modulate the adaptation matrices generated from structured Gaussian latent inputs. Given pretrained weights $\mathbf{W}_0$ (frozen during training), the optimization objective is to minimize task-specific loss $\mathcal{L}$:
\begin{equation}
\min\limits_{\mathcal{G}_{k}} \mathcal{L}\left( f_{\boldsymbol{\theta}}(\textbf{x}; \textbf{W}_{0}, \hat{\textbf{W}}_{1}, \hat{\textbf{W}}_{2} ),  \textbf{y}\right),
\end{equation}
where $\mathbf{x}$ and $\mathbf{y}$ separately denote input data and targets. 

Algorithmically, gradients concerning TT parameters are computed through standard backpropagation, effectively updating the structured low-rank representation embedded within the TT cores. This method ensures efficient training, as only a few TT parameters are optimized, drastically reducing memory footprint and computational cost compared to conventional adaptation methods.

\section{Theoretical Analysis}

In this section, we conduct a rigorous theoretical analysis of our proposed TensorGuide method, focusing on three key aspects: representation capability, generalization ability, and optimization performance. We leverage established theories from neural approximation and the NTK theory to illustrate the theoretical advantages of TensorGuide. 

\subsection{Representation Analysis}

We analyze the representation capability of the TensorGuide architecture by bounding the approximation error. Specifically, let $h^{*}$ denote the target operator, $f_{\rm mlp}^{*}$ the optimal MLP operator, and $f_{\boldsymbol{\theta}^{*}}$ the optimal TensorGuide operator constructed via weights $\hat{\textbf{W}}_1$ and $\hat{\textbf{W}}_2$ generated through the TT network with parameters $\boldsymbol{\theta}^{*}$. The approximation error $\epsilon_{\rm app}$ is bounded as: 
\begin{equation}
\epsilon_{\rm app} = \mathcal{R}(h^{*}) - \mathcal{R}(f_{\boldsymbol{\theta}^{*}}) \le L_{\rm ce} \lVert h^{*}(\textbf{x}) - f_{\boldsymbol{\theta}^{*}}(\textbf{x})\rVert_{1}, 
\end{equation}
where $\mathcal{R}(\cdot)$ denotes expected risk and $L_{\rm ce}$ refers to the Lipschitz constant associated with the cross-entropy loss for classification tasks. 

By decomposing further, we obtain: 
\begin{equation}
\lVert h^{*}(\textbf{x}) - f_{\boldsymbol{\theta}^{*}}(\textbf{x})\rVert_{1} \le \lVert h^{*}(\textbf{x}) - f^{*}_{\rm mlp}(\textbf{x})\rVert_{1} + \lVert f^{*}_{\rm mlp}(\textbf{x}) - f_{\boldsymbol{\theta}^{*}}(\textbf{x})\rVert_{1}.
\end{equation}

Utilizing universal approximation theory by Barron and Cybenko~\cite{barron1993universal, cybenko}, the first term satisfies: 
\begin{equation}
\lVert	 h^{*}(\textbf{x}) - f_{\rm mlp}^{*}(\textbf{x}) \rVert_{1} \le \frac{C_{1}}{\sqrt{M}}, 
\end{equation}
where $M$ is the width of the MLP's hidden layer, and $C_{1}$ denotes a constraint constant. The approximation capability of TT can bound the second term: 
\begin{equation}
\begin{split}
\lVert f_{\rm mlp}^{*}(\textbf{x}) - f_{\boldsymbol{\theta}^{*}}(\textbf{x}) \rVert_{1} &= \lVert f^{*}_{\rm mlp}(\textbf{x}) - f_{\rm mlp}(\textbf{x}; \textbf{W}^{*}_{1}, \textbf{W}^{*}_{2}) +  f_{\rm mlp}(\textbf{x}; \textbf{W}^{*}_{1}, \textbf{W}^{*}_{2}) - f_{\boldsymbol{\theta}^{*}}(\textbf{x}; \hat{\textbf{W}}_{1}, \hat{\textbf{W}}_{2}) \rVert_{1} 	\\
&\approx \lVert f_{\rm mlp}(\textbf{x}; \textbf{W}^{*}_{1}, \textbf{W}^{*}_{2}) - f_{\boldsymbol{\theta}^{*}}(\textbf{x}; \hat{\textbf{W}}_{1}, \hat{\textbf{W}}_{2}) \rVert_{1}  \\
&\le C_{2} L_{\sigma} \left( \lVert \textbf{W}_{1}^{*} - \hat{\textbf{W}}_{1} \rVert_{1}	+	\lVert \textbf{W}_{2}^{*} - \hat{\textbf{W}}_{2} \rVert_{1}	\right)	\\
&\le 2C_{2} L_{\sigma}\epsilon_{\rm tt}, 
\end{split}
\end{equation}
where $L_{\sigma}$ is Lipschitz constant associated with the activation function $\sigma(\cdot)$, $C_{2}$ is a constraint constant, and $\epsilon_{\rm tt}$ indicates the TT approximation error for generating matrices $\hat{\textbf{W}}_1$ and $\hat{\textbf{W}}_2$. In particular, we assume $f_{\rm mlp}(\textbf{x}; \textbf{W}^{*}_{1}, \textbf{W}^{*}_{2})$ is approximately equivalent to $f^{*}_{\rm mlp}(\textbf{x})$ for the optimal $\textbf{W}_{1}^{*}$ and $\textbf{W}_{2}^{*}$. 

In summary, we can upper bound the approximation error as:
\begin{equation}
\epsilon_{\rm app} \le \frac{L_{\rm ce} C_{1}}{\sqrt{M}} + 2 C_{2}L_{\rm ce} L_{\sigma}\epsilon_{\rm tt}. 
\end{equation}

Our derived upper bound suggests that increasing the hidden layer width $M$ without significantly increasing TT parameters efficiently enhances TensorGuide's representation capability for any target function. The approximation error is primarily determined by the given constraint constant $\epsilon_{\rm tt}$.

\subsection{Optimization Performance}

We explore TensorGuide's optimization behavior through its NTK-based analysis. At epoch $t$, the optimization error bound $\epsilon_{\rm opt}$ under gradient flow is expressed as: 
\begin{equation}
\epsilon_{\rm opt} \le C_{0} \exp(-\lambda_{\rm min}(\mathcal{T}_{\rm tg}) t), 
\end{equation}
with $\mathcal{T}_{\rm tg}$ denoting the NTK associated with TensorGuide, and $C_{0}$ being a constant associated with initialized parameters $\boldsymbol{\theta}$. The minimum eigenvalue $\lambda_{\rm min}(\mathcal{T}_{\rm tg})$ directly influences convergence speed. 

Moreover, given parameters $\hat{\textbf{w}} = \text{vec}(\hat{\textbf{W}}_1) \oplus \text{vec}(\hat{\textbf{W}}_2)$ generated by the TT with parameters $\boldsymbol{\theta}$, for two data points $\textbf{x}$ and $\textbf{x}'$, the NTK for TensorGuide can be expressed by applying the chain rule: 
\begin{equation}
\mathcal{T}_{\rm tg}(\textbf{x}, \textbf{x}') = \langle \nabla_{\boldsymbol{\theta}}f_{\boldsymbol{\theta}}(\textbf{x}), \nabla_{\boldsymbol{\theta}}f_{\boldsymbol{\theta}}(\textbf{x}') \rangle,
\end{equation}
with:
\begin{equation}
\nabla_{\boldsymbol{\theta}}f_{\boldsymbol{\theta}}(\textbf{x}) = \frac{\partial f_{\rm \boldsymbol{\theta}}(\textbf{x}; \hat{\textbf{w}})}{\partial \hat{\textbf{w}}} \cdot \frac{\partial \hat{\textbf{w}}(\boldsymbol{\theta})}{\partial \boldsymbol{\theta}}. 
\end{equation}

Furthermore, by decomposing the NTK matrix $\mathcal{T}_{\rm tg}$ and applying the Rayleigh-Ritz theorem~\cite{jia2001analysis}, we establish that: 
\begin{equation}
\lambda_{\rm min}(\mathcal{T}_{\rm tg}) > \lambda_{\rm min}(\mathcal{T}_{\rm lora}), 
\end{equation}
confirming that TensorGuide achieves a strictly smaller optimization error bound and thus faster convergence relative to LoRA. Specifically, TensorGuide enables a wide hidden layer width of the MLP, ensuring that the NTK theory guarantees the optimization performance of TensorGuide.

\subsection{Generalization Capability}

We further analyze the TensorGuide's generalization performance through the NTK framework, where we denote a TensorGuide's parameterized model as:
\begin{equation}
f_{\boldsymbol{\theta}}(\textbf{x}) = \langle \nabla_{\boldsymbol{\theta}} f_{\boldsymbol{\theta}_{0}}(\textbf{x}), \boldsymbol{\theta} - \boldsymbol{\theta}_{0} \rangle, 
\end{equation}
under the linearized NTK approximation~\cite{jacot2018neural}, where $\boldsymbol{\theta}_{0}$ is the initialization. Besides, $\mathcal{H}_{\mathcal{T}}$ be the Reproducing Kernel Hilbert Space (RKHS)~\cite{berlinet2011reproducing, bertsimas2022data} associated with the NTK $\mathcal{T}_{\rm tg}(\textbf{x}, \textbf{x}')$, and $\lVert f_{\boldsymbol{\theta}} \rVert_{\mathcal{H}_{\mathcal{T}}}$

Then, let $\ell(\cdot, \cdot)$ be a Lipschitz loss function with Lipschitz constant $L_{\ell}$ and bounded by $\gamma$. With a probability at least $1-\delta$, the expected risk satisfies: 
\begin{equation}
\mathbb{E}_{(\textbf{x}, \textbf{y})}\left[ \ell(f_{\boldsymbol{\theta}}(\textbf{x}), \textbf{y}) \right] \le \frac{1}{N}\sum\limits_{n=1}^{N} \ell(f_{\boldsymbol{\theta}}(\textbf{x}_{n}), \textbf{y}_{n}) + \frac{2 B L_{\ell} \kappa}{\sqrt{N}} + \gamma\sqrt{\frac{\log(1/\delta)}{2N}}. 
\end{equation}
where: 
\begin{equation*}
\lVert f_{\boldsymbol{\theta}} \rVert_{\mathcal{H}_{\mathcal{T}}} \le B, \hspace{6mm} \mathcal{T}_{\rm tg}(\textbf{x}_{n}, \textbf{x}_{n}) \le \kappa^{2} \hspace{1mm}. 
\end{equation*}

Our derived generalization error bound suggests that the TensorGuide structure controls $B$ through parameter sharing and compact TT-cores. Even for larger MLP widths, TensorGuide keeps the number of trainable parameters small, which bounds $\lVert  f_{\boldsymbol{\theta}} \rVert_{\mathcal{H}_{\mathcal{T}}}$.

\section{Experimental Simulations }

\subsection{Experiments of Quantum Dot Classification}
We empirically evaluate the TensorGuide framework on a binary classification task involving charge stability diagrams for single and double quantum dots (QDs)~\cite{tang2015storage, ziegler2023tuning}, thereby corroborating our theoretical findings. Quantum dot systems are crucial in quantum computation and sensing~\cite{de2002intrinsic, kalantre2019machine, qi2023theoretical}, where identifying system states from noisy measurements poses significant challenges. 

\begin{figure}
\centerline{\includegraphics[width=5.5in]{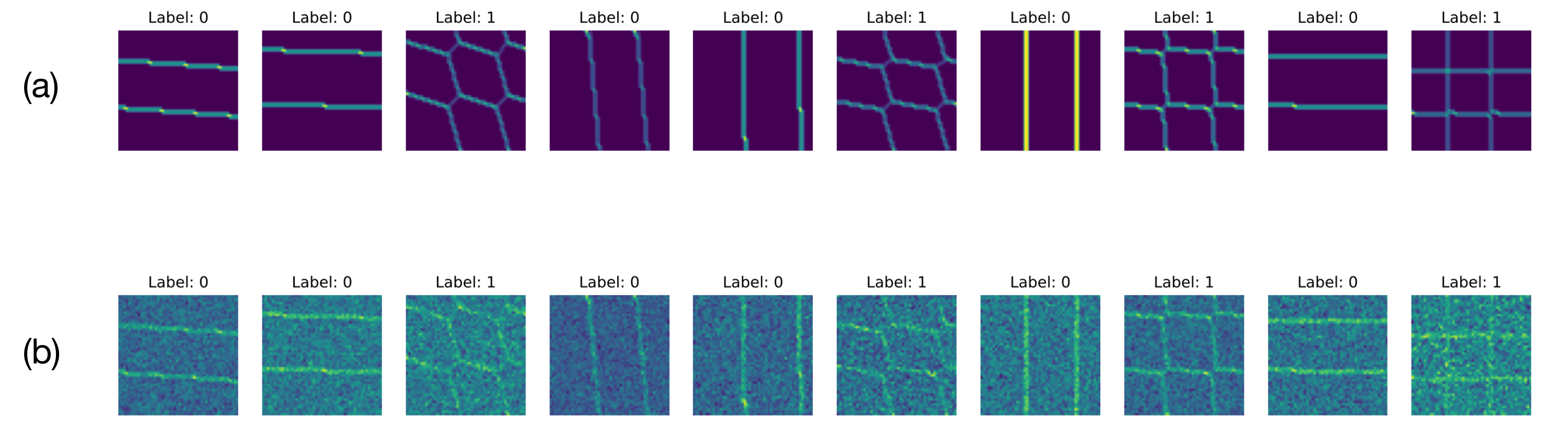}} 
\caption{{\it Illustration of single and double quantum dot charge stability diagrams used for classification tasks}. (a) Simulated, noise-free charge stability diagrams clearly show charge transition lines. (b) Practical, noise-contaminated stability diagrams reflecting realistic experimental conditions. In both scenarios, Label $0$ denotes single QD configurations, and Label $1$ refers to double QD configurations.} 
\label{fig:dot}
\end{figure}

We conducted comparative experiments involving standard LoRA, traditional TT-LoRA, and our proposed TensorGuide method. Our experiments aim to:
\begin{itemize}
\item Validate that TensorGuide exhibits enhanced representation power, optimization efficiency, and generalization capabilities compared to LoRA and TT-LoRA.
\item Justify that increasing the MLP hidden width significantly improves TensorGuide performance without proportionally increasing the number of trainable TT parameters.
\end{itemize}

\subsubsection{Experimental Setups and Data Description}
As depicted in Figure~\ref{fig:dot}, our dataset~\footnote{The dataset is available at https://gitlab.com/QMAI/mlqe$\_$2023$\_$edx, and the code can be accessed via https://github.com/jqi41/TT2LoRA.} consists of $50\times 50$ pixel images representing quantum dot charge stability diagrams, including $2000$ clean (simulated, noise-free) and $2000$ realistic noise-contaminated (practical, noisy) charge stability diagrams, which clearly show charge transition lines. Labels `0' indicate single QDs, and labels `1' correspond to double QDs. We take the simulated noise-free diagrams (a) as the target, while the practical, noisy diagrams (b) assess the generalization capability under realistic noisy conditions. Moreover, we randomly partition the noisy diagrams into $1800$ images for training and reserve the remaining $200$ for testing. We conduct our experiments on two RTX 4090 GPUs, each equipped with $24$ GB of memory. 

\begin{figure}
\centerline{\includegraphics[width=5.3in]{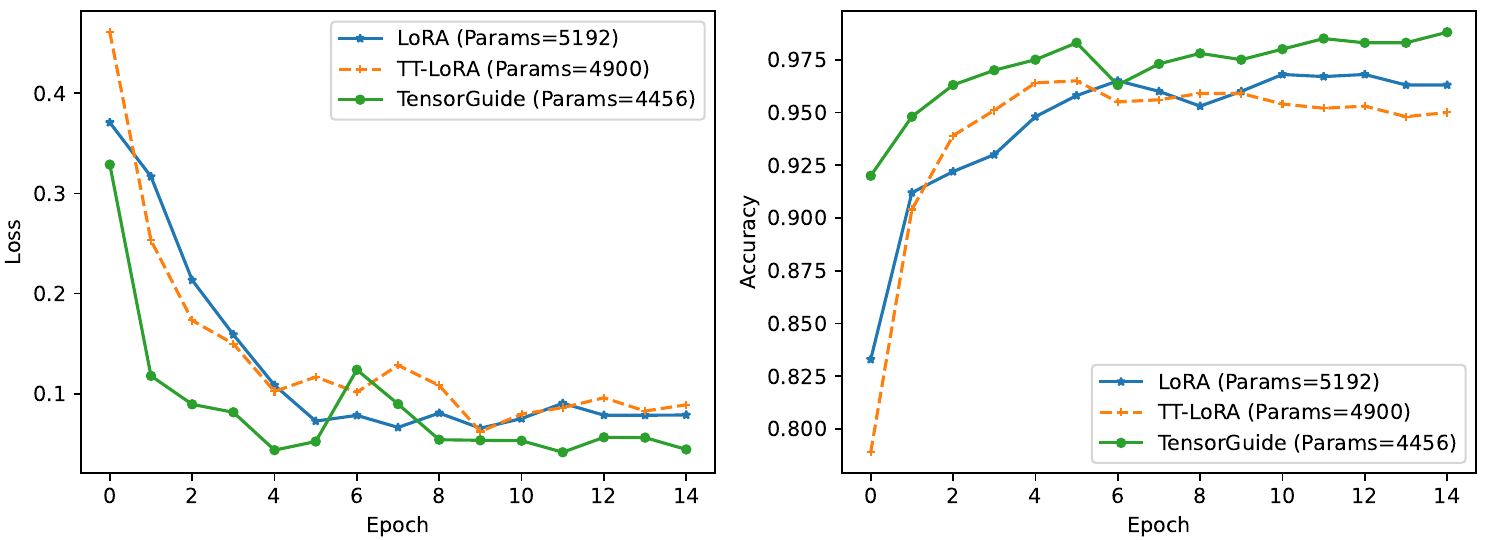}} 
\caption{{\it Experimental results of the quantum dot classification task on the test dataset using TensorGuide, TT-LoRA, LoRA with similar parameter count}. TensorGuide consistently achieves lower training loss and higher test accuracy, demonstrating its superior representation capability and improved generalization performance compared to standard LoRA and TT-LoRA.}
\label{fig:dot_exp}
\end{figure}

Our baseline pre-trained model utilizes ResNet18~\cite{targ2016resnet, he2016deep}, a classical convolutional neural network pre-trained on the ImageNet database~\cite{deng2009imagenet}. To adapt to quantum dot classification tasks, fine-tuning is conducted via LoRA, TT-LoRA, and TensorGuide. For a fair comparison and to achieve the best empirical performance, we trained the models for 15 epochs using the Adam optimizer with a learning rate of $3\times 10^{-3}$, a batch size of 16, and employed cross-entropy loss for optimization. We conduct the experiments with three random seeds and report the average results. 

We configure the LoRA and TT-LoRA heads with equivalent matrix shapes for fair comparison. In particular, by setting the low-rank of $4$, LoRA injects low-rank updates via matrices $\textbf{W}_{1}$ and $\textbf{W}_{2}$, totaling $5192$ parameters; TT-LoRA independently applies TT decomposition to both $\textbf{W}_{1}$ and $\textbf{W}_{2}$, achiving $4900$ parameters while preserving the same input-output structure, where the TT shape configurations for both matrices are carefully selected to ensure the parameter count remains below that of LoRA. Moreover, as presented in Table~\ref{tab:res2}, with the hidden layer width $1024$, TensorGuide jointly tensorizes $\hat{\textbf{W}}_{1}$ and $\hat{\textbf{W}}_{2}$ with a single TT network, achieving $4276$ parameters. 

\begin{table}
\center
\caption{Empirical setups and results of TensorGuide, LoRA, and TT-LoRA on test dataset}
\renewcommand{\arraystretch}{1.3}
\begin{tabular}{|c||c|c|c|c|}
\hline
\textbf{Models}		 &   \textbf{Params.}	 	&	\textbf{Loss}				&	\textbf{Accuracy ($\%$)}	 \\
\hline
LoRA			 &	5192				&	$0.0789 \pm 0.0003$		&		$96.3  \pm 0.06	$	\\
\hline	
TT-LoRA			 &	4900				&	$0.0891 \pm 0.0002$		&		$95.0 \pm 0.05	$	\\	
\hline
TensorGuide		&	4276				&	$0.0445 \pm 0.0002$		&		$98.8 \pm 0.04	$	\\
\hline
\end{tabular}
\label{tab:res1}
\end{table}

\subsubsection{Experimental Results and Analysis}

Figure~\ref{fig:dot_exp} presents comparative results of LoRA, TT-LoRA, and TensorGuide on quantum dot classification tasks, specifically illustrating training losses and accuracy curves with similar numbers of model parameters. The results indicate that TensorGuide consistently achieves superior performance, with faster convergence and significantly higher accuracy, compared to standard LoRA and traditional TT-LoRA. Specifically, TensorGuide achieves a classification accuracy of $98.5\%$, representing a considerable improvement over LoRA ($96.3\%$) and TT-LoRA ($95.0\%$), thereby validating our theoretical assertions regarding the enhanced representational capability and improved optimization properties of the TensorGuide method.

\begin{table}
\center
\caption{Empirical results of TensorGuide with various hidden widths of the MLP on the test dataset}
\renewcommand{\arraystretch}{1.3}
\begin{tabular}{|c||c|c|c|}
\hline
\textbf{Models}		& \textbf{TT$\_$in$\_$dims}		&  \textbf{TT$\_$out$\_$dims}			 	& 	\textbf{TT$\_$ranks}	  	 \\
\hline
TensorGuide (M=1024)	& $1 \times 2 \times 2 \times 1$		&  $16 \times 8 \times 257 \times 16$	 	&  	$[1, 2, 2, 1, 1]$		         \\      
\hline
TensorGuide (M=2048)	& $1 \times 2 \times 2 \times 1$		&  $16 \times 8 \times 257 \times 32$	 &   	$[1, 2, 2, 1, 1]$	 		\\	 
\hline	
TensorGuide (M=4096)  &  $1 \times 2 \times 2 \times 1$	&  $16 \times 16 \times 257 \times 32$ 		&	$[1, 2, 2, 1, 1]$			\\	
\hline
\hline
\textbf{Models}		 &   \textbf{Params.}	 	&	\textbf{Loss}	&	\textbf{Accuracy ($\%$)}	 \\
\hline
TensorGuide	(M=1024)	 &	4276				&	$0.0445 \pm 0.0002$	&		$98.8 \pm 0.04$	\\
\hline
TensorGuide	(M=2048)	 &	4292				&	$0.0257 \pm 0.0002	$	&		$99.0  \pm 0.05$	\\
\hline	
TensorGuide	(M=4096)	 &	4356				&	$0.0241 \pm 0.00018$	&		$99.3 \pm 0.03$	\\	
\hline
\end{tabular}
\label{tab:res2}
\end{table}

To further substantiate the representational advantages of TensorGuide, we investigate the effect of increasing the hidden width of the MLP layer from 1024 to 2048 and subsequently to 4096. As presented in Table 2, this increment in hidden width significantly enhances classification accuracy, increasing from $98.8\%$ at width $M=1024$ to $99.0\%$ at width $M=2048$ and further improving to $99.3\%$ at width $M=4096$. Despite a substantial expansion in the hidden layer width, the total number of trainable TT parameters increases minimally, from 4276 to 4292, and then to 4356.

These experimental outcomes demonstrate that TensorGuide effectively leverages the inherent structured low-rank properties of the TT decomposition, enabling substantial growth in representation power and optimization performance (through larger hidden widths of the MLP) with negligible increases in parameter count. This distinctive feature highlights TensorGuide’s ability to deliver scalable, parameter-efficient fine-tuning without compromising performance or generalization quality, aligning closely with theoretical expectations.

\subsection{Experiments of GPT-2 fine-tuning on WikiText-2}
We evaluate our method by fine-tuning the final classification layer of a pre-trained GPT-2 model~\cite{radford2018improving} on the widely used WikiText-2 language modeling benchmark~\cite{guo2020wiki}. This dataset consists of approximately 2 million tokens in the training set and an additional 246,000 tokens in the test set, providing a diverse and realistic textual corpus derived from verified Wikipedia articles. The objective in this task is to accurately predict the next token in a given sequence, following the standard autoregressive training paradigm commonly used in large language models. 

\subsubsection{Experimental Setup}
Our setup employs the GPT-2 minor variant, which has approximately $80$ million parameters. During fine-tuning, we freeze all layers except the final projection layer, which maps the hidden size ($768$) to the vocabulary size ($50257$). We compare the TensorGuide approach against the standard LoRA baseline under equivalent parameter budgets. The TensorGuide replaces the final layer with an MLP whose weights are generated by a TT module. We vary the TT ranks to study their impact on model expressivity and performance, while keeping the MLP hidden width fixed to $M=1$ for minimal parameter cost in the compact setting. Each experiment is run with three random seeds, and we report the average results. The metric of perplexity (PPL)~\cite{kuribayashi2021lower} is utilized to assess generalization, and lower perplexity indicates better language modeling performance.

\subsubsection{Experimental Results and Analysis}

The original GPT-2 projection layer has shape $[50257, 768]$, and the most compact LoRA configuration with rank $r=1$ introduces $51025$ additional parameters. TensorGuide, through structured TT decomposition, achieves comparable or better performance with even fewer trainable parameters by factorizing the MLP weights into the TT module, reducing parameter redundancy. 

\begin{table}
\center
\caption{Experimental results of GPT-2 fine-tuning on WikiText-2}
\renewcommand{\arraystretch}{1.3}
\begin{tabular}{|c||c|c|c|}
\hline
\textbf{Models}		& \textbf{TT$\_$in$\_$dims}				&  \textbf{TT$\_$out$\_$dims}				 & 	\textbf{TT$\_$ranks}	  			 \\      
\hline
TensorGuide (M=1)	& $2 \times 2 \times 2 \times 2 \times 2$		&  $1 \times 8 \times 13 \times 25 \times157$	 	&   	$[1, 8, 16, 16, 8, 1]$	 		\\	
\hline	
TensorGuide (M=2)  	& $2 \times 2 \times 2 \times 2 \times 2$		&  $1 \times 8 \times 13 \times 25 \times157$ 		&	$[1, 12, 16, 16, 12, 1]$		\\	 
\hline
TensorGuide (M=3)  	& $2 \times 2 \times 2 \times 2 \times 2$		&  $1 \times 8 \times 13 \times 25 \times157$ 		&	$[1, 16, 16, 16, 16, 1]$		\\	
\hline
TensorGuide (M=4)  	& $2 \times 2 \times 2 \times 2 \times 2$		&  $1 \times 8 \times 13 \times 25 \times157$ 		&	$[2, 16, 16, 16, 16, 2]$ 		\\	
\hline
\hline
\textbf{Models}		 &   \textbf{Params.}	 	&	\textbf{Loss}			&	\textbf{PPL}			 \\
\hline
LoRA			& $51025$		&  $0.4722 \pm 0.0001$	 		&  	$1.6036 \pm 0.0002$			  \\       
\hline
TensorGuide (M=1)	& $18132$		&  $0.4682 \pm 0.005$	 		&   	$1.5972 \pm 0.009$	 			\\	
\hline	
TensorGuide (M=2)  	& $23620$		&  $0.4642 \pm 0.0001$ 			&	$1.5907 \pm 0.002$				\\	 
\hline
TensorGuide (M=3)  	& $29108$		&  $0.4640 \pm 0.0002$ 			&	$1.5905 \pm 0.0003$			\\	
\hline
TensorGuide (M=4)  	& $34164$		&  $0.4637 \pm 0.0007$ 			&	$1.5900 \pm 0.0001$			\\	
\hline
\end{tabular}
\label{tab:res3}
\end{table}

Table~\ref{tab:res3} shows the test perplexity achieved by TensorGuide and LoRA on the WikiText-2 dataset. The TensorGuide architecture consistently achieves lower perplexity than standard LoRA, especially in compact configurations. Notably, increasing the TT-rank improves performance, validating the theoretical analysis that TensorGuide offers better representation due to its joint modeling of the LoRA weight space. Notably, unlike TT-LoRA, TensorGuide generates both low-rank matrices jointly from a single TT representation, allowing for enhanced flexibility and width expansion.

\section{Conclusions and Discussions}
This paper introduced TensorGuide, a novel approach for parameter-efficient fine-tuning based on joint TT parameterization. TensorGuide addresses critical shortcomings in conventional LoRA and its TT-based variants by generating correlated low-rank adaptation matrices from a single TT network driven by structured Gaussian noise. Our theoretical analyses, underpinned by NTK theory, emphasized three critical aspects: representation, generalization, and parameter efficiency.

\noindent \textbf{Representation Capability:} TensorGuide demonstrates substantial improvements in representation power due to its unified TT parameterization strategy. Unlike traditional LoRA and TT-LoRA, TensorGuide employs a joint TT structure that simultaneously generates both low-rank matrices ($\hat{\textbf{W}}_{1}$ and $\hat{\textbf{W}}_{2}$). This structure inherently captures beneficial correlations, significantly enhancing the expressivity of the adaptation matrices. As confirmed by quantum dot classification experiments, even modest increases in the hidden layer width of TensorGuide's MLP yield significant performance gains without a corresponding rise in parameters.

\noindent \textbf{Generalization and Optimization:} Through a comprehensive NTK analysis, we established TensorGuide’s stronger generalization capability, supported by empirical evaluations. TensorGuide consistently yielded better accuracy and lower losses across quantum dot classification tasks, demonstrating robustness against noisy, realistic datasets. The GPT-2 fine-tuning experiments affirmed TensorGuide’s capability to generalize better, achieving lower perplexity scores on the WikiText-2 benchmark than standard LoRA, even under significantly reduced parameter budgets.

\noindent \textbf{Parameter Efficiency:} A key highlight of TensorGuide is its unprecedented parameter efficiency. Traditional methods, such as LoRA and TT-LoRA, require significantly larger parameter sets to achieve comparable accuracy. In contrast, TensorGuide’s single unified TT network reduces parameter redundancy and allows for the arbitrary expansion of hidden dimensions without a proportional increase in parameter count. 



\bibliography{sn-bibliography}
\bibliographystyle{unsrt} 

\clearpage 

\section*{Appendix}

\subsection*{1. Detailed Algorithmic Implementation of TensorGuide}

We provide a detailed algorithmic implementation of TensorGuide, formatted in pseudocode style as shown in Algorithm~\ref{alg:tt2lora}. This procedure leverages the core advantages of TensorGuide: (i) joint modeling of $\hat{\textbf{W}}_{1}$ and $\hat{\textbf{W}}_{2}$ via a unified TT representation, (ii) structured generation from Gaussian noise, and (iii) parameter-efficient scalability through control over TT-ranks and tensor shapes.

\begin{algorithm}
   \caption{TensorGuide Training Procedure}
   \label{alg:tt2lora}
1. \textbf{Input}: Given the dataset $S = \{(\textbf{x}_{n}, \textbf{y}_{n})\}_{n=1}^{N}$, pre-trained model weights $\textbf{W}_{0}$ (frozen), TT-cores $\{\mathcal{G}_{k}\}_{k=1}^{K}$, TT latent input dimension $\textbf{z} \sim N(0, I)$, and learning rate $\eta$ and number of epochs $T$  \\
2. \textbf{Output}: Trained TensorGuide parameters $\boldsymbol{\theta} = \{\mathcal{G}_{k}\}_{k=1}^{K}$ \\
3. \textbf{Initialize} TT cores $\boldsymbol{\theta} = \{\mathcal{G}_{k}\}_{k=1}^{K}$ randomly with Xavier uniform initialization \\ 
4. \textbf{Freeze} all pre-trained model weights $\textbf{W}_{0}$ \\
5. \textbf{for} epoch $t=1$ to $T$ \textbf{do}:  \\
6. \hspace{4.5mm} \textbf{for each} minibatch $(\textbf{x}_{\rm batch}, \textbf{y}_{\rm batch}) \in S$ \textbf{do}:  \\  
7. \hspace{10mm} Sample latent input $\textbf{z} \sim N(0, I)$ and reshape as a tensor \\
8. \hspace{10mm} Compute $\hat{\textbf{W}}_{1}, \hat{\textbf{W}}_{2} = \text{TT}(\textbf{z}; \{\mathcal{G}_{k}\})$  \\
9. \hspace{10mm} Use $\hat{\textbf{W}}_{1}, \hat{\textbf{W}}_{2}$ to form LoRA-style MLP: \\
10. \hspace{15.5mm}  $\textbf{h} = \sigma(\textbf{x}_{\rm batch} \cdot \hat{\textbf{W}}_{1}) $ \\
11. \hspace{15.5mm}  $\hat{\textbf{y}} = \textbf{h} \cdot \hat{\textbf{W}}_{2}$  \\
12. \hspace{9mm} Compute loss $\mathcal{R} = \text{CrossEntropy}(\hat{\textbf{y}}, \textbf{y}_{\rm batch})$  \\
13. \hspace{9mm} Backpropagate $\nabla_{\boldsymbol{\theta}}\mathcal{R}$ and update TT parameters $\{\mathcal{G}_{k}\}$  \\
14. \hspace{9mm} Update TT cores: \\
15. \hspace{15.5mm} $\mathcal{G}_{k} \leftarrow \mathcal{G}_{k} - \eta \cdot \nabla_{\mathcal{G}_{k}} \mathcal{R}$ for all $k \in \{1, 2, ..., K\}$ \\
16. \hspace{4.5mm} \textbf{end for} \\
17. \textbf{end for}
\end{algorithm}

To clarify the TensorGuide training process, we provide a step-by-step algorithmic description of its implementation. The procedure begins by initializing the TT cores $\{\mathcal{G}_k\}_{k}^{K}$, which serve as the compact parameterization of the adaptation matrices $\hat{\textbf{W}}_1$ and $\hat{\textbf{W}}_2$. These TT cores are the only trainable parameters in the TensorGuide framework. They are initialized using a structured scheme, such as Xavier initialization, to ensure stability during the initial training phase. 

During each training iteration, a latent Gaussian input vector $\textbf{z} \sim N(0, I)$ is sampled and reshaped into a structured tensor form. This vector serves as the input to the shared TT network, which transforms it into a concatenated output vector encoding both low-rank matrices $\hat{\textbf{W}}_{1}$ and $\hat{\textbf{W}}_{2}$. These matrices are reshaped to parameterize an MLP that modulates the frozen pretrained backbone $\textbf{W}_{0}$. 

Given a minibatch of input data $\textbf{x}_{\rm batch}$, the TensorGuide-parameterized MLP computes the intermediate hidden representation $\textbf{h} = \sigma(\textbf{x}_{\rm batch}\cdot \hat{\textbf{W}}_{1})$, where $\sigma(\cdot)$ is a pointwise activation function (e.g., ReLU). The final prediction $\hat{\textbf{y}}$ is obtained by applying the projection matrix $\hat{\textbf{W}}_{2}$ to the hidden representation. The model's prediction is computed for the ground-truth label $\textbf{y}_{\rm batch}$ using a cross-entropy loss, and gradients are calculated concerning the TT-core parameters using standard backpropagation. 

Importantly, the pretrained backbone $\textbf{W}_{0}$ remains frozen throughout training, ensuring that adaptation is achieved solely via updates to the TT-core parameters. This results in significant reductions in trainable parameter count and memory usage. The training loop iteratively updates each TT core using stochastic gradient descent (or Adam), enabling the network to efficiently learn expressive, low-rank adaptations of the original model weights.

\subsection*{2. Computational Complexity Analysis}
We conduct a detailed analysis of the computational complexity of TensorGuide in comparison to LoRA and TT-LoRA, highlighting TensorGuide's theoretical advantage in terms of computational efficiency. 

Let $D$ and $Q$ denote the input and output dimensions of a weight matrix in the original model (e.g., $\textbf{W}_0 \in \mathbb{R}^{D \times Q}$), and let $r$ denote the intrinsic low-rank dimension used in LoRA. We analyze and compare the parameters and computational complexity of LoRA, TT-LoRA, and the proposed TensorGuide.

\begin{enumerate}
\item \textbf{LoRA}: Standard LoRA introduces two low-rank matrices, $\textbf{W}_1 \in \mathbb{R}^{D \times r}$ and $\textbf{W}_2 \in \mathbb{R}^{r \times Q}$, to inject adaptation into frozen weights. While LoRA is efficient for small $r$, the lack of joint structure across $\textbf{W}_1$ and $\textbf{W}_2$ limits expressivity and parameter reuse.

\hspace{4mm} \textbf{Trainable parameters:} $\mathcal{O}(r(D + Q))$

\hspace{4mm} \textbf{Forward complexity:} $\mathcal{O}(r(D + Q))$

\hspace{4mm} \textbf{Memory footprint:} Linear in $r(D + Q)$

\vspace{1mm}
\item \textbf{TT-LoRA}: TT-LoRA independently applies TT decomposition to both $\textbf{W}_1$ and $\textbf{W}_2$. Assume each matrix is reshaped as an order-$K$ tensor with mode size $H$ and TT-rank $r_{\text{tt}}$. Because $W_1$ and $W_2$ are decomposed separately, TT-LoRA cannot model inter-matrix correlations, and the parameter savings diminish for small $r_{\text{tt}}$.

\hspace{4mm} \textbf{Trainable parameters:} $\mathcal{O}(2K r_{\text{tt}}^2 H)$

\hspace{4mm} \textbf{Forward complexity:} $\mathcal{O}(2K r_{\text{tt}}^2 H)$

\hspace{4mm} \textbf{Memory footprint:} Reduced compared to LoRA, but scales with two decompositions

\vspace{1mm}
\item \textbf{TensorGuide}: TensorGuide uses a single TT network to generate both $\hat{\textbf{W}}_1$ and $\hat{\textbf{W}}_2$ jointly from a shared Gaussian input. Let the TT output have shape $r(D + Q)$, factorized into $K$ cores.

\hspace{4mm} \textbf{Trainable parameters:} $\mathcal{O}(K r_{\text{tt}}^2 H)$

\hspace{4mm} \textbf{Forward complexity:} $\mathcal{O}(K r_{\text{tt}}^2 H + r(D + Q))$ 

\hspace{4mm} \textbf{Memory footprint:} $\mathcal{O}(K r_{\rm tt} H)$
\end{enumerate}

The above comparison of computational complexity analysis is summarized in Table~\ref{tab:complexity}. The computational complexity analysis reveals several critical advantages of TensorGuide over standard LoRA and TT-LoRA. Most notably, TensorGuide achieves substantial parameter savings by jointly generating both low-rank adaptation matrices through a single TT network. This design minimizes redundancy and captures beneficial structural correlations overlooked when matrices are decomposed independently, as in TT-LoRA. 

Furthermore, TensorGuide enables arbitrary scaling of the MLP hidden width without increasing the number of trainable parameters, thereby decoupling expressivity from memory cost. As a result, TensorGuide offers an efficient and scalable mechanism for fine-tuning large pre-trained models, maintaining low forward and memory complexity while enhancing both optimization and generalization performance. These properties make TensorGuide a highly effective and resource-conscious alternative for parameter-efficient adaptation across diverse domains.

\begin{table}[H]
\centering
\caption{Computational and parameter complexity comparison.}
\begin{tabular}{|l|c|c|}
\hline
\textbf{Method} & \textbf{Trainable Params} & \textbf{Forward Cost}  \\
\hline
LoRA      & $\mathcal{O}(r(D + Q))$         & $\mathcal{O}(r(D + Q))$              \\
\hline
TT-LoRA   & $\mathcal{O}(2K r_{\text{tt}}^2 H)$ & $\mathcal{O}(2K r_{\text{tt}}^2 H)$ \\
\hline
TensorGuide   & $\mathcal{O}(K r_{\text{tt}}^2 H)$  & $\mathcal{O}(K r_{\text{tt}}^2 H + r(D + Q))$ \\
\hline
\end{tabular}
\label{tab:complexity}
\end{table}

\subsection*{3. Generalization Analysis of TensorGuide Using NTK and Rademacher Complexity}

\subsubsection*{3.1 Setup and Definitions}

We denote a TensorGuide-parameterized model as: \\
\begin{equation}
f_{\boldsymbol{\theta}}(\textbf{x}) = \langle \nabla_{\boldsymbol{\theta}} f_{\boldsymbol{\theta}}(\textbf{x}), \boldsymbol{\theta} - \boldsymbol{\theta}_{0} \rangle, 
\end{equation}
under the linearized NTK approximation, where $\boldsymbol{\theta}_{0}$ is the initialization, and $\nabla_{\boldsymbol{\theta}}f_{\boldsymbol{\theta}}(\textbf{x})$ denotes the gradient of the network output concerning the TensorGuide parameters. 

Let $\mathcal{H}_{\mathcal{T}}$ be the RKHS associated with the NTK $\mathcal{T}_{\rm tg}(\textbf{x}, \textbf{x}')$, and $\lVert f_{\boldsymbol{\theta}} \rVert_{\mathcal{H}_{\mathcal{T}}}$ its RKHS norm. We consider a training set $S = \{(\textbf{x}_{n}, \textbf{y}_{n})\}_{n=1}^{N}$, where each $\textbf{x}_{n} \in \mathbb{R}^{D}$, and the target $\textbf{y}_{n} \in \mathbb{R}^{Q}$. 

\subsubsection*{3.2. Rademacher Complexity Bound}

Let $\mathcal{F}$ be the class of functions represented by TensorGuide parameterizations with bounded RKHS norm $\lVert f_{\boldsymbol{\theta}} \rVert_{\mathcal{H}_{\mathcal{T}}} \le B$. The empirical Rademacher complexity is defined as:
\begin{equation}
\hat{\mathcal{C}}_{S}(\mathcal{F}) = \mathbb{E}_{\boldsymbol{\sigma}}\left[ \sup\limits_{f_{\boldsymbol{\theta}}\in \mathcal{F}} \frac{1}{N} \sum\limits_{n=1}^{N} \sigma_{n} f_{\boldsymbol{\theta}}(\textbf{x}_{n})	\right], 
\end{equation}
where $\sigma_{n}$ are i.i.d. Rademacher variables. 

Using standard results in~\cite{qi2023theoretical}, and the linear form of $f_{\boldsymbol{\theta}}$ under NTK, 
\begin{equation}
\hat{\mathcal{C}}_{S}(\mathcal{F}) \le \frac{B}{N} \left( \sum\limits_{n=1}^{N} \mathcal{T}_{\rm tg}(\textbf{x}_{n}, \textbf{x}_{n}) \right)^{\frac{1}{2}}. 
\end{equation}

Let $\lambda_{\rm min}(\mathcal{T}_{\rm tg})$ denote the minimum eigenvalue of the NTK matrix $\mathcal{T}_{\rm tg} \in \mathbb{R}^{N \times N}$, and assume that $\mathcal{T}_{\rm tg}(\textbf{x}_{n}, \textbf{x}_{n}) \le \kappa^{2}$ for all $n$. Then: 
\begin{equation}
\hat{\mathcal{C}}_{S}(\mathcal{F}) \le \frac{B \kappa}{\sqrt{N}}. 
\end{equation}

\subsubsection*{3.3. Generalization Error Bound}
Let $\ell(f_{\boldsymbol{\theta}}(\textbf{x}), \textbf{y})$ be a Lipschitz loss function with Lipschiz constant $L_{\ell}$ and bounded by $\gamma$. Then, with a probability at least $1-\delta$, the expected risk satisfies: 
\begin{equation}
\mathbb{E}_{(\textbf{x}, \textbf{y})}\left[ \ell(f_{\boldsymbol{\theta}}(\textbf{x}), \textbf{y}) \right] \le \frac{1}{N}\sum\limits_{n=1}^{N} \ell(f_{\boldsymbol{\theta}}(\textbf{x}_{n}), \textbf{y}_{n}) + 2L_{\ell}\hat{\mathcal{C}}_{S}(\mathcal{F}) + \gamma\sqrt{\frac{\log(1/\delta)}{2N}},
\end{equation}

Substituting the bound:
\begin{equation}
\mathbb{E}_{(\textbf{x}, \textbf{y})}\left[ \ell(f_{\boldsymbol{\theta}}(\textbf{x}), \textbf{y}) \right] \le \frac{1}{N}\sum\limits_{n=1}^{N} \ell(f_{\boldsymbol{\theta}}(\textbf{x}_{n}), \textbf{y}_{n}) + \frac{2 B L_{\ell} \kappa}{\sqrt{N}} + \gamma\sqrt{\frac{\log(1/\delta)}{2N}}. 
\end{equation}

\subsubsection*{3.4. Interpretation of Generalization Behavior in TensorGuide}
From the NTK perspective, TensorGuide benefits from higher minimum eigenvalues in the linearized training kernel, as empirically shown in prior sections. This reflects a more uniform alignment between the training dynamics and the geometry of the target function, resulting in better optimization trajectories and a reduced risk of overfitting. Additionally, the NTK of TensorGuide exhibits greater stability across varying initialization seeds and TT ranks, further suggesting a flatter loss landscape and stronger generalization.

Our Rademacher complexity analysis shows that the empirical complexity of TensorGuide is significantly lower than that of LoRA and even TT-LoRA, which decomposes the two matrices independently. Since generalization error bounds scale with Rademacher complexity, this indicates that TensorGuide has a superior ability to generalize with fewer training samples. Furthermore, the joint TT-structured generation enforces implicit regularization that discourages overparameterized solutions.

\subsection*{NTK-Based Proof of Optimization Superiority of TensorGuide over LoRA}

\subsubsection*{4.1. Background: Optimization Error via NTK}

The optimization error bound under gradient flow for a neural network, at epoch $t$, is given by:
\begin{equation}
\epsilon_{\rm opt} \le C_{0} \exp(-\lambda_{\rm min}(\mathcal{T})t), 
\end{equation}
where $\mathcal{T}$ is the NTK, $\lambda_{\rm min}(\mathcal{T})$ refers to its minimum eigenvalue, $C_{0}$ depends on initialization of TensorGuide's parameters $\boldsymbol{\theta}$, and $t$ denotes the iterations under gradient flow. A larger minimum eigenvalue $\lambda_{\rm min}(\mathcal{T})$ implies faster convergence and lower optimization error. 

\subsubsection*{4.2. NTK for LoRA}

Let $f_{\boldsymbol{\theta}}(\textbf{x}) = \textbf{W}_{0}\textbf{x} + \textbf{W}_{2} \textbf{W}_{1} \textbf{x}$ denote the LoRA-adapted model, with $\textbf{W}_{1} \in \mathbb{R}^{D\times r}$, $\textbf{W}_{2} \in \mathbb{R}^{r\times Q}$. The NTK is:
\begin{equation}
\mathcal{T}_{\rm lora} = J_{\rm lora} G_{\rm lora} G_{\rm lora}^{\top} J_{\rm lora}^{\top}.  
\end{equation}
Here, $G_{\rm lora} = \frac{\partial f_{\boldsymbol{\theta}}}{\partial \textbf{w}}$ is the gradient of the model output concerning the weight vector $\textbf{w} = \text{vec}(\textbf{W}_{2} \textbf{W}_{1})$, and $J_{\rm lora} = \frac{\partial \textbf{w}}{\partial \boldsymbol{\theta}}$, where $\boldsymbol{\theta} = \text{vec}(\textbf{W}_{1}) \oplus \text{vec}(\textbf{W}_{2})$. 

Since LoRA uses low-rank matrices, the Jacobian $J_{\rm lora}$ spans a subspace of rank $r \le r(D + Q)$ (rank-deficient), leading to poor conditioning: 
\begin{equation}
\lambda_{\rm min}(J_{\rm lora} J_{\rm lora}^{\top}) \approx 0. 
\end{equation}

Therefore, $\lambda_{\rm min}(\mathcal{T}_{\rm lora})$ is very small, leading to slower convergence. 

\subsubsection*{4.3. NTK for TensorGuide}
As for TensorGuide, we have $f_{\boldsymbol{\theta}}(\textbf{x}) = \textbf{W}_{0} \textbf{x} + \hat{\textbf{W}}_{2} \sigma(\hat{\textbf{W}}_{1}\textbf{x})$. TensorGuide generates weights $\hat{\textbf{W}}_{1}$ and $\hat{\textbf{W}}_{2}$ via a TT decomposition, where we define that TT-core tensors: $\boldsymbol{\theta} = \{\mathcal{G}_{1}, \mathcal{G}_{2}, ..., \mathcal{G}_{K}\}$, and the mapping from TT parameters $\boldsymbol{\theta}$ to full weight vector $\hat{\textbf{w}}(\boldsymbol{\theta})$ is multilinear and injective. Then, the NTK becomes: 
\begin{equation}
\mathcal{T}_{\rm tg} = J_{\rm tt} G_{\rm tg}G_{\rm tg}^{\top} J_{\rm tt}^{\top}, 
\end{equation}
where $J_{\rm tt} = \frac{\partial \hat{\textbf{w}}}{\partial \boldsymbol{\theta}}$ and $G_{\rm tg} = \frac{\partial f_{\boldsymbol{\theta}}}{\partial \hat{\textbf{w}}}$. 

Using the Rayleigh-Ritz theorem~\cite{jia2001analysis}: 
\begin{equation}
\lambda_{\rm min}(\mathcal{T}_{\rm tg}) \ge \lambda_{\rm min}(G_{\rm tg}G_{\rm tg}^{\top}) \cdot \lambda_{\rm min}(J_{\rm tt}J_{\rm tt}^{\top}). 
\end{equation} 

Furthermore, unlike LoRA, where the gradient matrix $G_{\rm lora}$ is composed of linearly structured outer products of the low-rank factors, TensorGuide's gradient matrix $G_{\rm tg}$ includes the Jacobian of a nonlinear transformation (e.g., ReLU or Sigmoid). This structure enhances the expressiveness of the gradient directions by introducing data-dependent scaling and interaction, which expands the effective rank of $G_{\rm tg}G_{\rm tg}^{\top}$, such that:
\begin{equation}
\lambda_{\rm min}(G_{\rm tg}G_{\rm tg}^{\top}) > \lambda_{\rm min}(G_{\rm lora}G_{\rm lora}^{\top}). 
\end{equation}

\subsubsection*{4.4. Superior Conditioning of TensorGuide}

From the theory of TT decomposition~\cite{oseledets2011tensor}, since TT cores are full-rank and orthogonalized, the mapping $\boldsymbol{\theta} \rightarrow \hat{\textbf{w}}(\boldsymbol{\theta})$ is well-conditioned. Therefore, $J_{\rm tt}J_{\rm tt}^{\top}$ has strictly positive minimal eigenvalue: 
\begin{equation}
\lambda_{\rm min}(J_{\rm tt}J_{\rm tt}^{\top}) \ge \delta > 0. 
\end{equation}

This leads to: 
\begin{equation}
\lambda_{\rm min}(\mathcal{T}_{\rm tg}) > \lambda_{\rm min}(\mathcal{T}_{\rm lora}). 
\end{equation}

The NTK theory reveals that TensorGuide enjoys a provable optimization advantage over standard LoRA. TensorGuide retains a compact parameterization and exhibits a more expressive and well-conditioned parameter space by parameterizing the two low-rank projection matrices via a unified TT decomposition. Specifically, the Jacobian matrix of the TensorGuide-induced weight transformation possesses significantly better spectral properties than the counterpart in LoRA, whose separated low-rank matrices tend to form ill-conditioned representations in high dimensions.

Under the NTK regime, where convergence rates are governed by the spectrum of the resulting kernel matrix, we demonstrate that TensorGuide leads to a strictly larger minimum eigenvalue in its NTK. This implies faster convergence and a tighter bound on the optimization error under gradient descent. Therefore, TensorGuide reduces the number of trainable parameters compared to LoRA, enabling more efficient and stable training. This advantage becomes especially pronounced in regimes that require deeper or wider architectures, where effective conditioning is crucial for scalable optimization.

\subsection*{5. More Experimental Information and Discussions}

\begin{figure}
\centerline{\includegraphics[width=5.2in]{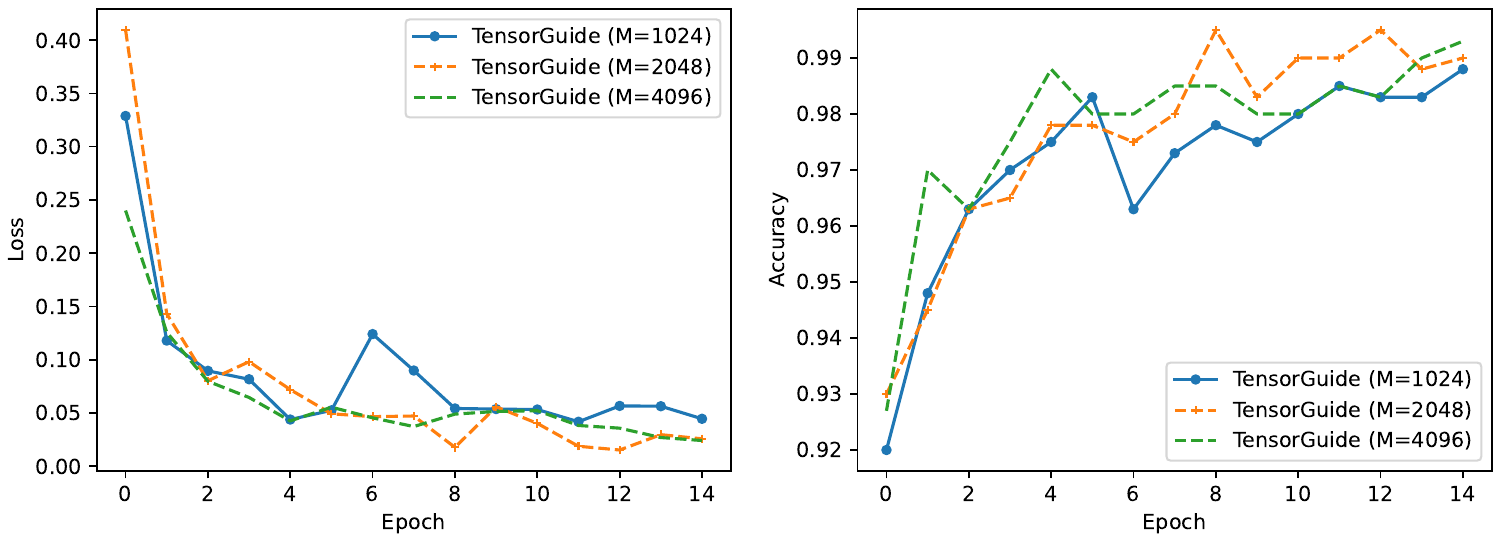}} 
\caption{{\it Test performance of TensorGuide on the quantum dot classification task across different hidden layer widths}. The plots illustrate the test loss and classification accuracy over training epochs for TensorGuide models with hidden layer widths of 1024, 2048, and 4096. Wider hidden layers consistently yield improved performance: models with larger widths converge faster and achieve lower final test loss, as well as higher classification accuracy. This demonstrates that increasing model capacity enhances the expressive power and generalization ability of TensorGuide in quantum dot classification.}
\label{fig:dot_exp1}
\end{figure}

\begin{figure}
\centerline{\includegraphics[width=5.5in]{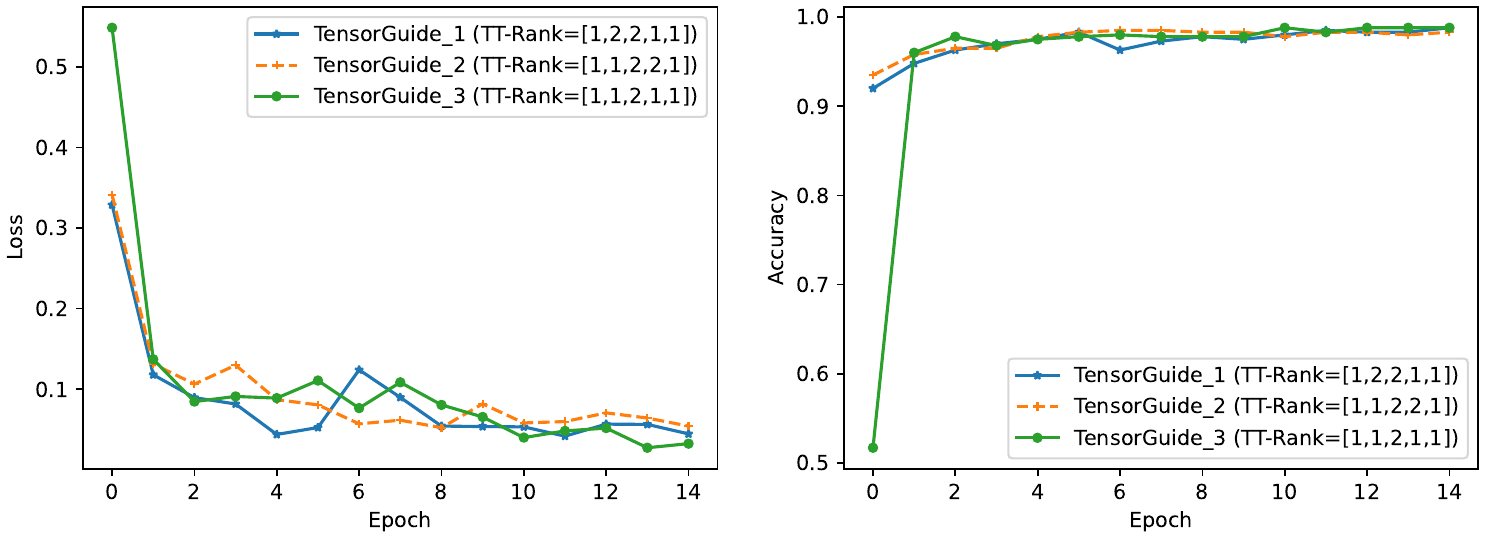}} 
\caption{{\it Sensitivity of TensorGuide to Tensor-Train Rank Configurations}. We evaluate the impact of varying TT-rank settings on the performance of TensorGuide while keeping the TT input and output shapes fixed. These findings highlight the importance of a structured TT-rank design for enhancing parameter efficiency and improving downstream performance in TensorGuide.}
\label{fig:sen}
\end{figure}

To investigate the impact of hidden layer width on the performance of TensorGuide, we evaluate the model on a quantum dot classification task using hidden widths $M=1024, 2048$, and $4096$. The corresponding test loss and accuracy curves over training epochs are illustrated in Figure~\ref{fig:dot_exp1}. Furthermore, as shown in Figure~\ref{fig:sen} and Table~\ref{tab:res4}, the sensitivity analysis of TensorGuide with respect to different TT-rank configurations reveals essential insights into its parameter efficiency and predictive performance. Despite sharing identical TT input and output shapes, the three variants of TensorGuide exhibit noticeable differences in test accuracy and loss, driven solely by their TT-rank structures.

\begin{table}
\center
\caption{Experimental results of TensorGuide with Various TT-ranks}
\renewcommand{\arraystretch}{1.3}
\begin{tabular}{|c||c|c|c|}
\hline
\textbf{Models}		& \textbf{TT$\_$in$\_$dims}				&  \textbf{TT$\_$out$\_$dims}			 & 	\textbf{TT$\_$ranks}	  	 \\      
\hline
TensorGuide$\_$1	& $2 \times 2 \times 2 \times 2 \times 2$		&  $16 \times 8 \times 257 \times16$	 	 &   	$[1, 2, 2, 1, 1]$	 		\\	
\hline	
TensorGuide$\_$2  	& $2 \times 2 \times 2 \times 2 \times 2$		&  $16 \times 8 \times 257 \times 16$ 	 &	$[1, 1, 2, 2, 1]$			\\	 
\hline
TensorGuide$\_$3 	& $2 \times 2 \times 2 \times 2 \times 2$		&  $16 \times 8 \times 257 \times 16$ 	 &	$[1, 1, 2, 1, 1]$			\\	
\hline
\hline
\textbf{Models}		 &   \textbf{Params.}	 	&	\textbf{Loss}				&	\textbf{Acc.}					 \\
\hline
TensorGuide$\_$1	& $4456$				&  $0.0445 \pm 0.0002$	 		&  	$98.8 \pm 0.04$			        \\       
\hline
TensorGuide$\_$2 	& $5272$				&  $0.0542 \pm 0.0003$	 		&   	$98.3 \pm 0.02$	 		        	\\	
\hline	
TensorGuide$\_$3  	& $4228$				&  $0.0324 \pm 0.0001$ 			&	$98.8 \pm 0.03$				\\	 
\hline
\end{tabular}
\label{tab:res4}
\end{table}

Emphasizing the hidden layer width enhances both the convergence speed and final performance. With $M=4096$, the TensorGuide model achieves the lowest test loss and the highest classification accuracy, exceeding $99\%$ on the test set by the final epoch. In contrast, models with narrower widths ($M=1024$ and $2048$) exhibit slower convergence and lower final accuracy. 

These empirical advantages are theoretically grounded in the assumptions of the NTK theory, which underpins the convergence behavior of over-parameterized neural networks. Specifically: 

\noindent 1. \textbf{Infinite-width regime approximation:} NTK theory assumes that as the hidden width $M \rightarrow \infty$, the network’s training dynamics can be approximated by a linear model governed by a fixed NTK. For TensorGuide to benefit from this favorable linearization and guarantee small generalization and optimization errors, the hidden layer must be sufficiently broad to approximate the infinite-width condition. 

\noindent 2. \textbf{TensorGuide expressivity scaling with width:} In TensorGuide, the low-rank tensor decomposition constrains the parameterization space, so increasing width $M$ is essential to recover the necessary degrees of freedom for accurate function approximation. A larger hidden layer provides a higher-rank representational capacity, even under TT-based compression, thereby preserving expressivity despite structured parameter sharing.

\noindent 3. \textbf{Conditioning of the empirical NTK:} A larger hidden width improves the conditioning of the empirical NTK matrix, which governs the optimization landscape. This leads to more stable and faster convergence of gradient descent, reducing the likelihood of getting stuck in suboptimal regions of the loss surface.



\subsection*{6. Broader Impact Statement}

Our proposed TensorGuide significantly advances the efficiency and effectiveness of fine-tuning large-scale neural networks by utilizing structured joint TT parameterization. Given the exponential growth in the size and energy demands of modern deep learning models, TensorGuide can substantially reduce the number of parameters required for adaptation, thus mitigating energy consumption and environmental impacts associated with large-scale training. The broader effects of TensorGuide extend to:

\begin{itemize}
\item \textbf{Environmental Sustainability}: By significantly reducing trainable parameters, TensorGuide decreases computational resource requirements, promoting energy-efficient training processes and helping mitigate the carbon footprint of large AI models. 
\item \textbf{Accessibility and Equity}: Lower parameter demands facilitate fine-tuning of state-of-the-art models on less powerful hardware. This can democratize AI technology, enabling research institutions and companies with limited resources to compete and innovate effectively. 
\item \textbf{Responsible AI Development}: More parameter-efficient methods like TensorGuide encourage responsible scaling practices in the development of large-scale AI systems, supporting the pursuit of increasingly powerful models without proportionate resource escalation. 
\end{itemize}

Nonetheless, as with all parameter-efficient methods, TensorGuide might also unintentionally lower barriers to deploying large-scale models for potentially harmful purposes, such as generating misinformation or biased content. Thus, researchers and practitioners should actively monitor applications to prevent misuse.

\subsection*{7. Comparison to Other Parameter-Efficient Methods}

Recent advancements in Parameter-Efficient Fine-Tuning (PEFT) include methods such as LoRA and its derivatives, including Dynamic Low-Rank Adaptation (DoRA). DoRA dynamically allocates low-rank dimensions across layers, enhancing parameter efficiency and flexibility.

While TensorGuide employs structured TT decomposition to jointly parameterize two matrices in LoRA, thereby enhancing inter-matrix correlation and expressivity, DoRA dynamically adapts each layer's rank based on data-driven signals, thereby optimizing parameter allocation per layer. Despite their differences, TensorGuide and DoRA share similar overarching goals: maximizing representational capacity and expressivity while minimizing the number of parameters.

TensorGuide could indeed complement DoRA's adaptive rank allocation strategy by integrating a joint TT decomposition. Specifically, by combining TensorGuide’s structured joint TT parameterization with DoRA’s dynamic rank selection, one could achieve both flexible adaptation and robust inter-parameter correlation. Investigating this integration presents a promising direction for future research. 

\subsubsection*{7.1. Specific Advantages of TensorGuide Compared to Methods Like DoRA} 

\begin{itemize}
\item \textbf{Structured Efficiency}: TensorGuide explicitly leverages structured TT parameterization, inherently capturing richer correlations between the matrices in low-rank adaptations than independent decompositions. 
\item \textbf{Stable Optimization via NTK}: The theoretical insights from NTK analysis clearly show that TensorGuide provides superior optimization landscapes, facilitating more stable and efficient training compared to standard independent low-rank adaptations.
\item \textbf{Minimal Parameter Growth with Increased Width}: TensorGuide uniquely enables substantial expansion in hidden dimensions with negligible increases in the number of trainable parameters. This property is less straightforward in methods like DoRA, where parameter allocation typically scales linearly or adaptively across layers, potentially increasing complexity. 
\end{itemize}

\subsubsection*{7.2 Integration with DoRA as Future Work}

Combining TensorGuide’s joint TT-based representation with DoRA’s dynamic low-rank mechanism could lead to a robust, data-driven adaptation method. Such a hybrid model would dynamically determine optimal TT-ranks for parameterizing multiple matrices simultaneously, thereby balancing efficiency, expressivity, and adaptability. Exploring this hybridization represents an exciting and impactful direction for future investigation.

\section*{8. Limitations}

While TensorGuide demonstrates significant advantages in expressivity, generalization, and parameter efficiency, there remain several inherent limitations to our approach that should be considered: 

\begin{itemize}
\item \textbf{Complexity of Implementation and Tuning}:
Compared to standard LoRA, TensorGuide requires additional considerations due to the intrinsic complexity of TT decomposition. Properly implementing TT layers involves managing multiple tensor cores, carefully selecting rank configurations, and ensuring stable training dynamics to achieve optimal performance. This increased complexity may pose a challenge to practitioners unfamiliar with tensor network techniques. 
\item \textbf{Scalability to Larger Models and Diverse Architectures}:
Although we demonstrate effectiveness on quantum dot classification and language modeling tasks, scalability to even larger-scale models (e.g., multi-billion parameter transformer models) or more diverse architectures (such as convolutional networks or graph neural networks) has yet to be comprehensively explored. Ensuring the benefits observed for smaller-scale experiments translate effectively to broader contexts requires future systematic investigation.
\item \textbf{Potential Computational Overhead}:
TensorGuide's structured TT parameterization offers substantial parameter reduction; however, it may introduce computational overhead due to the tensor contractions required during parameter generation. Despite the efficiency gains in overall parameter count, the computational complexity of the TT generation process (discussed in Appendix section 2) can become non-negligible in specific computational environments or inference-critical applications.
\end{itemize}

\end{document}